# TESTING FOR CAUSALITY IN CONTINUOUS TIME BAYESIAN NETWORK MODELS OF HIGH-FREQUENCY DATA

JONAS HALLGREN,*† *Royal Institute of Technology, KTH*

TIMO KOSKI,* *Royal Institute of Technology, KTH*

**Abstract**

Continuous time Bayesian networks are investigated with a special focus on their ability to express causality. A framework is presented for doing inference in these networks. The central contributions are a representation of the intensity matrices for the networks and the introduction of a causality measure. A new model for high-frequency financial data is presented. It is calibrated to market data and by the new causality measure it performs better than older models.

                                         Secondary 65S99

## 1. Introduction

Continuous time Bayesian networks (CTBNs) are graphical representations of the dependence structures between continuous time random processes with finite state spaces. The name stems from a certain similarity to Bayesian networks, which give a graphical representation between random variables, rather than from shared mathematical properties. CTBNs were introduced by Nodelman et al. (2002) independently of Schweder (1970), who called them Composable processes. In this paper the name CTBN will primarily be used. An important feature of the CTBNs is their ability to express causality. In time series analysis, a process has a causal effect if other time series are more precisely predicted given the causing process. This concept was introduced by Granger (1969). Schweder's Composable processes can be seen as a continuous time version of Granger's causality. The link is also discussed in Didelez (2007). Florens and Fougere (1996) define different concepts in noncausality in continuous time and charcterize these in the case of Markov processes; thus extending Schweder (1970). Aalen et al. (2012) discusses Schweder's theory from a causality and martingale point of view. Gourieroux et al. (1987) have the same approach to causality as we do; they allow general state spaces but are limited to discrete time. Gégout-Petit and Commenges (2010) are also working from the martingale perspective and introduces a theory of influence between stochastic processes. Eichler (2012) considers composability in a time series setting. Sugihara et al. (2012) perform inference using the causal time series perspective in an eco-system.

* Postal address: Department of Mathematics, Royal Institute of Technology, KTH, SE-100 44 Stockholm, Sweden
† Email address: jonas@math.kth.se

Directed information theory is linked conceptually and theoretically to Granger causality, see Amblard and Michel (2012). Hlaváčková-Schindler et al. (2007) explore the information theoretic concepts for detecting causality in time series. The directed information is defined for the information transfered from one instant to another in discrete time. Kaiser and Schreiber (2002) consider the directed information for processes in discrete time and takes the limit, thereby forming a continuum of instances. A similar approach is taken by Weissman et al. (2013) who define a directed information measure in continuous time by taking the limit of the discrete time directed information measure. These papers have an approach to causality in continuous time that is similar to ours–although distinctively different. Seghouane and Amari (2012) have an approach that is similar to the directed information theory but with terminology close to ours. Causality for continuous time Bayesian networks is defined with the continuous time processes as a starting point without a limiting procedure. The causality measure is then defined directly for the process as opposed to taking the limit of a measure for a discrete process.

The object of this paper is to develop tools that simplify inference and detect

---

Hoem (1969) cites a working paper on Composable processes by Schweder from 1968, predating Granger's paper. However, Schweder refrain from explicitly mentioning causality.

causality in the CTBNs. Let $W$ be a continuous time stochastic process with two components, $(X,Y)$, taking values in the finite space $\widetilde{\mathsf{W}} = \mathsf{X} \times \mathsf{Y}$. The process $W$ is called a CTBN or a *Composable finite Markov process* if it satisfies

$$\lim_{h \to 0} \frac{1}{h} \mathbb{P}(X_{t+h} \neq x, Y_{t+h} \neq y \mid X_t = x, Y_t = y) = 0. \tag{1}$$

That is, for a sufficiently small interval, the probability that more than one of the two component processes has changed their state tends to zero. The processes $X$ and $Y$ are called *composable* if $W = (X,Y)$ satisfies the composable property above. Throughout the paper the composable process will have two components. The contributions of the paper are as follows. CTBNs are Composable finite Markov processes and vice versa. The intensity matrix of a CTBN must have a specific structure. This structure is described. We design an operator which takes components and composes them into a CTBN. We provide an ordering for this composition which simplifies inference. A new model for tick-by-tick financial data is proposed and calibrated using the tools developed in the paper. A new measure of causality is proposed and demonstrated on a simulated CTBN.

The paper consists of seven sections. In this introduction some background and the definition of a CTBN is given. In Section 2 a key lemma is derived. The lemma lays the foundation for the construction of an operator in Section 3 which describes how CTBNs are represented as Markov processes. An ordering of the states and an example illustrating the operator are presented. Section 4 introduces the Kullback–Leibler divergence for two components in a CTBN. The divergence is used to define the causal relation between two components in a network. Section 5 applies the methodology developed to a novel model for financial tick data. In Section 6 the financial model is calibrated. A simulation study is done on the model from Section 3 demonstrating the frameworks ability to detect causality. The paper is concluded with a discussion.

## 2. Graphical representation

Let $\{W_t\}_{t \geq 0}$ be a CTBN with state space $\mathsf{W}$. Let $\{X_t\}_{t \geq 0}$ and $\{Y_t\}_{t \geq 0}$ be continuous time stochastic processes with state spaces $\mathsf{X}$ and $\mathsf{Y}$ respectively. Throughout the paper our objects of interest are finite state spaces. Thus, without loss of generality, assume that $\mathsf{X} = \{1, 2, \ldots, n^X\}$ and $\mathsf{Y} = \{1, 2, \ldots, n^Y\}$. Let $W$ take values in $\mathsf{W} = \{1, 2, \ldots, n^W\}$ where there is a one-to-one correspondence from $\mathsf{X} \times \mathsf{Y} = \widetilde{\mathsf{W}}$ to $\mathsf{W}$ implying that $n^W = n^X \cdot n^Y$. Details on the correspondence are given in Section 3.2. In this paper all Markov processes are assumed to be time homogeneous. Elements in an intensity matrix for a time homogeneous continuous time Markov process $W_t$ are defined as

$$q(w_j \mid w_i) = \mathbb{Q}_{ij}^W = \lim_{h \to 0} \frac{1}{h}\mathbb{P}(W_{t+h} = w_j \mid W_t = w_i),$$

$$q(w_i \mid w_i) = \mathbb{Q}_{ii}^W = -\sum_{j}^{n^W} \mathbb{Q}_{ij}^W,$$

and are constant in $t$, see Doob (1953) for details. Inspired by this, define for $X$

$$q(x_j \mid x_i, y_k) = \lim_{h \to 0} \frac{1}{h}\mathbb{P}(X_{t+h} = x_j \mid X_t = x_i, Y_t = y_k), \quad (2)$$

and an analogous function for $Y$. In the next section an operator producing intensity matrices for CTBNs is presented. The construction of that operator is based on the following lemma.

**Lemma 1.** *Let $W = (X, Y)$, be a CTBN. Then the intensity for $W$ is given by*

$$q(x_j, y_\ell \mid x_i, y_k) = \begin{cases} q(x_j \mid x_i, y_k), & x_j \neq x_i, \\ q(y_\ell \mid x_i, y_k), & y_\ell \neq y_k, \\ q(x_i \mid x_i, y_k) + q(y_k \mid x_i, y_k), & x_j = x_i, y_\ell = y_k, \\ 0, & x_j \neq x_i, y_\ell \neq y_k. \end{cases} \quad (3)$$

The proof is given in Appendix A.

**2.1. Local Independence**

The process $X$ is said to be *locally independent* (LI) of $Y$ if $q(x_i \mid x_j, y)$ is constant in $y$. If $W = (X, Y)$ is a composable process and $X$ is locally independent of $Y$ then $X$ is a Markov process since $q(x_i \mid x_j, y) = q(x_i \mid x_j)$ makes up the elements of an intensity matrix $\mathbb{Q}^X$ for $X$. This was proved by Schweder (1970), and with the formulation provided by Lemma 1 it is evident.

That $X$ is locally independent of $Y$ is equivalent to that $X_{t+h} \mid X_t$ is independent of $Y_t$, as proved by Schweder (1970). In other words the values of $X_{t+h}$ are not affected by $Y_t$ conditioned that $X_t$ is known. This is denoted by $X \not\leftarrow Y$. Lack of local independence can be interpreted as presence of causality. Thus there are four possible combinations of causality: $X \leftrightarrow Y$, or $X \not\rightarrow Y$ and $Y \leftarrow X$, or $X \leftarrow Y$ and $Y \not\rightarrow X$, or $X \not\rightarrow Y$ and $Y \not\rightarrow X$, see Figure 1. A result from Schweder (1970) states that $X$ and $Y$ are stochastically independent if and only if they are both locally independent of each other. That is, the first three configurations in Figure 1 must have dependence between $X$ and $Y$ while there can be none in the last one.

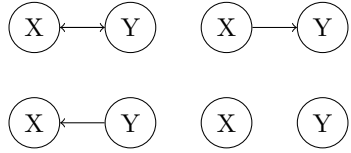

FIGURE 1: Four possible combinations of causal relationships

## 3. Composing intensity matrices

The object of this section is to compute the intensity matrix for a CTBN $W = (X, Y)$. If the process $X_{t+h} \mid Y_t = y_k$ is Markov, which is not necessarily the case, define a *conditional Markov process* with transition probabilities $q(x_i \mid x_j, y_k)$. These quantities make up the *conditional intensity matrix* $\mathbb{Q}^{X|y_k}$.

### 3.1. Conditional Markov Processes

Assume that the process $X_t \mid Y_t = y_k$ is a continuous time Markov process for all $y_k$ and ditto for $Y_t \mid X_t = x_i$. The intensity matrix for $X$ is denoted by $\mathbb{Q}^{X|y_k}$ and an element $\mathbb{Q}^{X|y_k}_{ij}$ of the intensity matrix is given by $q(x_j \mid x_i, y_k)$ as defined in Equation (2). By Lemma 1 it is possible to construct $\mathbb{Q}^W$ using the matrices $\mathbb{Q}^{X|y_k}$ and $\mathbb{Q}^{Y|x_i}$ for $k = 1, \ldots, n^Y, i = 1, \ldots, n^X$. An operation which arranges the conditional intensity matrices in a suitable way is therefore desired. The particular choice of operation presented here will utilize full intensity matrices of sizes $n^X n^Y \times n^X n^Y$ denoted by $\mathbb{Q}^{X|Y}$ and $\mathbb{Q}^{Y|X}$ respectively. These matrices will be specific arrangements of all the elements from the conditional intensity matrices. Lemma 1 implies that the intensity matrices must be arranged such that they interfere only on the main diagonal; any other arrangement would violate the lemma.

A natural way to organize the intensity matrix for $X \mid Y$ is to divide it into $n^Y$

blocks of size $n^X \times n^X$

$$\begin{bmatrix} \mathbb{Q}^{X|y_1} & 0 & 0 & 0 \\ 0 & \mathbb{Q}^{X|y_2} & 0 & 0 \\ 0 & 0 & \ddots & 0 \\ 0 & 0 & 0 & \mathbb{Q}^{X|y_{n^Y}} \end{bmatrix}, \qquad (4)$$

this is known as the matrix direct sum and is denoted $\bigoplus_{k=1}^{n^Y} \mathbb{Q}^{X|y_k}$. By definition, if $X$ is locally independent of $Y$ then the blocks are all identical. On the other hand, if the blocks are all identical then the function $q(x_j \mid x_i, y)$ is constant in $y$ and, again by definition, $X \not\leftarrow Y$.

For the operator working on $\mathbb{Q}^{Y|X}$ the Kronecker product is involved. Let $B$ and $A$ be be two quadratic matrices of size $n_B$ and $n_A$ respectively. Recall that the Kronecker product between the matrix $B$ and any other matrix $C$ is defined as

$$B \otimes C = \begin{bmatrix} b_{11} C & \cdots & b_{1n_B} C \\ \vdots & \ddots & \vdots \\ b_{n_B 1} C & \cdots & b_{n_B n_B} C \end{bmatrix}.$$

Thus, for two quadratic matrices $B$ and $C$, of size $n_B$ and $n_C$ respectively, their Kronecker product $B \otimes C$ is of size $n_B n_C \times n_B n_C$. Define

$$\bigotimes_{k=1}^{n_A} B_k = \sum_{k=1}^{n} B_k \otimes I_k^{[n_A]}. \qquad (5)$$

The new operator fulfills the specifications above; it arranges the conditional intensity matrices into a full intensity matrix such that the elements on the off-diagonals will not collide with the ones from the direct sum. The new operator takes several matrices $B_k$ (of size $n_B \times n_B$), extends them to a larger space of size $n_B n_A \times n_B n_A$ and then sums them together. This is similar to the Kronecker sum. Let $b_{ij}^{(k)}$ denote element $i, j$ of the matrix $B_k$. Then the operation gives the $n_B n_A \times n_B n_A$ matrix

$$\bigotimes_{k=1}^{n_A} B_k = \begin{bmatrix} b_{11}^{(1)} & 0 & \cdots & 0 & b_{12}^{(1)} & 0 & \cdots & 0 & \cdots \\ 0 & b_{11}^{(2)} & \cdots & 0 & 0 & b_{12}^{(2)} & \cdots & 0 & \cdots \\ \vdots & 0 & \ddots & 0 & \vdots & 0 & \ddots & 0 & \cdots \\ 0 & \cdots & 0 & b_{11}^{(n_A)} & 0 & \cdots & 0 & b_{12}^{(n_A)} & \cdots \\ b_{21}^{(1)} & 0 & \cdots & 0 & b_{22}^{(1)} & 0 & \cdots & 0 & \cdots \\ 0 & b_{21}^{(2)} & \cdots & 0 & 0 & b_{22}^{(2)} & \cdots & 0 & \cdots \\ \vdots & 0 & \ddots & 0 & \vdots & 0 & \ddots & 0 & \cdots \\ 0 & \cdots & 0 & b_{21}^{(n_A)} & 0 & \cdots & 0 & b_{22}^{(n_A)} & \cdots \\ 0 & \cdots & 0 & 0 & \ddots & \cdots & 0 & 0 & \ddots \end{bmatrix}.$$

Define the $n_A \times n_A$ matrix

$$\tilde{B}_{jk} = \begin{bmatrix} b_{jk}^{(1)} & 0 & \cdots & 0 \\ 0 & b_{jk}^{(2)} & \cdots & 0 \\ \vdots & 0 & \ddots & 0 \\ 0 & \cdots & 0 & b_{jk}^{(n_A)} \end{bmatrix},$$

then we can write

$$\bigotimes_{k=1}^{n_A} B_k = \begin{bmatrix} \tilde{B}_{11} & \tilde{B}_{12} & \cdots & \tilde{B}_{1n_B} \\ \tilde{B}_{21} & \tilde{B}_{22} & \cdots & \tilde{B}_{2n_B} \\ \vdots & \cdots & \ddots & \vdots \\ \tilde{B}_{n_B 1} & \tilde{B}_{n_B 2} & \cdots & \tilde{B}_{n_B n_B} \end{bmatrix}.$$

The direct sum from Equation (4) is defined as

$$\bigoplus_{k=1}^{n_B} A_k = \begin{bmatrix} A_1 & 0 & 0 & 0 \\ 0 & A_2 & 0 & 0 \\ 0 & 0 & \ddots & 0 \\ 0 & 0 & 0 & A_{n_B} \end{bmatrix}.$$

As a result, in the operation $\bigoplus_{k=1}^{n_B} A_k + \bigotimes_{k=1}^{n_A} B_k$, it is clear that only diagonal elements of the matrices $A_k$ and $B_k$ collides. The purpose of the operators is to provide us with a way to compose several single processes in to one. We end up with the following proposition.

**Proposition 1.** *Let $W = (X, Y)$, be a process where $X \mid Y$ and $Y \mid X$ are continuous time Markov processes. Then, given the conditional matrices split up by state, the intensity matrix from Lemma 1 can be written as*

$$\mathbb{Q}^W = \bigoplus_{k=1}^{n_Y} \mathbb{Q}^{X|y_k} + \bigotimes_{k=1}^{n_X} \mathbb{Q}^{Y|x_k}, \tag{6}$$

*if and only if $X$ and $Y$ are composable.*

The operation in (6) is not unique. Arranging the state space of the CTBN in a different manner requires the operation to change. Nevertheless, it should still produce something equivalent. An example of an intensity matrix for a CTBN is seen in Figure 2. The dark diagonal elements are negative, the positive elements are light and the remaining elements are zero-valued. The figure shows two processes, one with a state space of cardinality ten and the other with five. On the main diagonal there are five blocks, $\mathbb{Q}^{X|y_k}$, of size $10 \times 10$, one for each possible state of the second process. The other intensity matrix, $\mathbb{Q}^{Y|x_i}$, is spread out as the diagonal bands in the matrix.

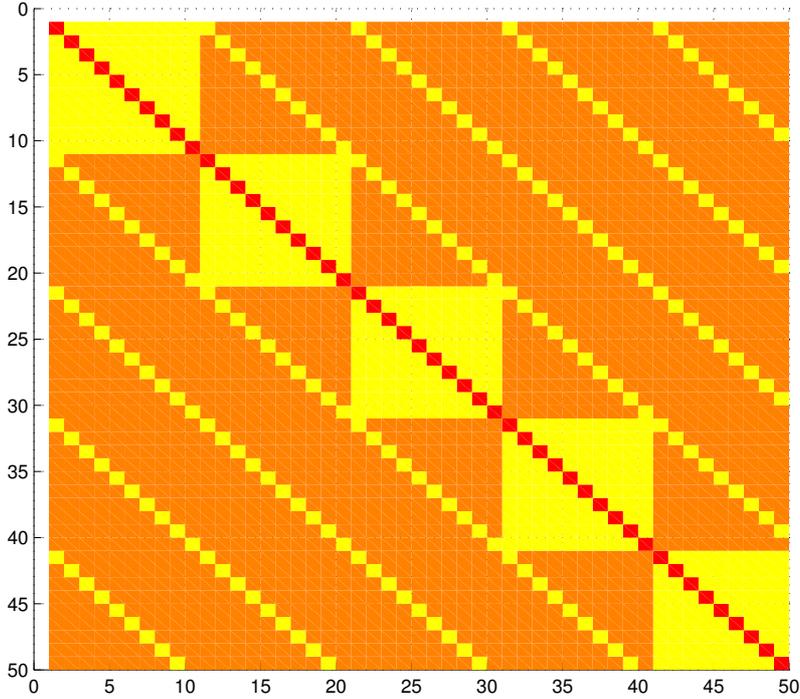

Figure 2: Intensity matrix for a $10 \times 5$ Composable process

### 3.2. Ordering of the States

The ordering of the states is essential. The operation $\bigoplus_{k=1}^{n_Y} \mathbb{Q}^{X|y_k} + \bigotimes_{k=1}^{n_X} \mathbb{Q}^{Y|x_k}$ produces one particular ordering. We regard this as a natural one. It is induced by an ordering of X and Y. For integer-valued $X$ and $Y$ the ordering is self-evident. When the states are of the type "roe" and "fish" the ordering is arbitrary and must be decided.

The ordering produced by $\bigoplus_{k=1}^{n^Y} \mathbb{Q}^{X|y_k} + \bigotimes_{k=1}^{n^X} \mathbb{Q}^{Y|x_k}$ can be written as

$$\mathsf{W} = \{1, 2, \ldots, n^X, n^X + 1, \ldots, n^X + n^X(2-1), \ldots, n^X n^Y\}. \tag{7}$$

The intensity matrix for $\mathbb{Q}^W$ will have $n^Y$ blocks of size $n^X \times n^X n^Y$ given as

$$\begin{bmatrix} x \mid y_1 \\ x \mid y_2 \\ \vdots \\ x \mid y_n \end{bmatrix}.$$

This means that each row of the intensity matrix $\mathbb{Q}^W$ will have the ordering given by (7). For $X$ and $Y$ this corresponds to the states

$$\widetilde{\mathsf{W}} = \{\{1,1\}, \ldots, \{n^X, 1\}, \ldots, \{1, n^Y\}, \ldots, \{n^X, n^Y\}\}.$$

This state space is perhaps more easily understood as the matrix

$$\begin{bmatrix} 1,1 & 2,1 & \cdots & n^X, 1 \\ 1,2 & 2,2 & \cdots & n^X, 2 \\ \vdots & \vdots & \ddots & \vdots \\ 1, n^Y & \cdots & \cdots & n^X, n^Y \end{bmatrix},$$

where each increase in column represents an increase of state in $X$ and each row the same for $Y$. The corresponding numbering of the states is given by

$$\begin{bmatrix} 1 & 2 & \cdots & n^X \\ n^X + 1 & n^X + 2 & \cdots & n^X + n^X(2-1) \\ \vdots & \vdots & \ddots & \vdots \\ 1 + n^X(n^Y - 1) & \cdots & \cdots & n^X n^Y \end{bmatrix}.$$

We end up with an ordering that we can compute from the relations given in the following proposition where $\lceil \cdot \rceil$ denotes the ceiling function.

**Proposition 2.** *For three states $w$, $x$ and, $y$ of $W, X$ and $Y$ respectively we have the relations*

$$\begin{aligned} w &= x + n^X(y-1), \\ x &= 1 + (w - 1 \mod n^X), \\ y &= \left\lceil \frac{w}{n^X} \right\rceil. \end{aligned} \tag{8}$$

These relations depend on the operations in $\bigoplus_{k=1}^{n_Y} \mathbb{Q}^{X|y_k} + \bigotimes_{k=1}^{n_X} \mathbb{Q}^{Y|x_k}$ and might be invalidated if the operators are defined in a different manner.

### 3.3. Example: Markov Process with Markov Modulated Intensity

Consider a process $X$ switching between two states $\{1,2\}$ with intensities given by a process $Y$ taking values in $\{1,2\}$, that is,

$$\mathbb{Q}^{X|y_1} = \begin{bmatrix} -\lambda^{(1)} & \lambda^{(1)} \\ \mu^{(1)} & -\mu^{(1)} \end{bmatrix}, \qquad \mathbb{Q}^{X|y_2} = \begin{bmatrix} -\lambda^{(2)} & \lambda^{(2)} \\ \mu^{(2)} & -\mu^{(2)} \end{bmatrix}.$$

The process $Y$ will have similar matrices,

$$\mathbb{Q}^{Y|x_1} = \begin{bmatrix} -\beta^{(1)} & \beta^{(1)} \\ \gamma^{(1)} & -\gamma^{(1)} \end{bmatrix}, \qquad \mathbb{Q}^{Y|x_2} = \begin{bmatrix} -\beta^{(2)} & \beta^{(2)} \\ \gamma^{(2)} & -\gamma^{(2)} \end{bmatrix}.$$

Further, let the processes be composable. The intensity matrix for the CTBN is given by Proposition 1 as

$$\bigoplus_{k=1}^{2} \mathbb{Q}^{X|y_k} + \bigotimes_{k=1}^{2} \mathbb{Q}^{Y|x_k} = \begin{bmatrix} -\lambda^{(1)}-\beta^{(1)} & \lambda^{(1)} & \beta^{(1)} & 0 \\ \mu^{(1)} & -\mu^{(1)}-\beta^{(2)} & 0 & \beta^{(2)} \\ \gamma^{(1)} & 0 & -\lambda^{(2)}-\gamma^{(1)} & \lambda^{(2)} \\ 0 & \gamma^{(2)} & \mu^{(2)} & -\mu^{(2)}-\gamma^{(2)} \end{bmatrix}$$

with state space $\{\{1,1\},\{2,1\},\{1,2\},\{2,2\}\}$. We can write this matrix in several equivalent ways but note that doing so implies that we rearrange the order of the state space. For instance,

$$\bigoplus_{k=1}^{2} \mathbb{Q}^{Y|x_k} + \bigotimes_{k=1}^{2} \mathbb{Q}^{X|y_k} = \begin{bmatrix} -\lambda^{(1)}-\beta^{(1)} & \beta^{(1)} & \lambda^{(1)} & 0 \\ \gamma^{(1)} & -\lambda^{(2)}-\gamma^{(1)} & 0 & \lambda^{(2)} \\ \mu^{(1)} & 0 & -\mu^{(1)}-\beta^{(2)} & \beta^{(2)} \\ 0 & \mu^{(2)} & \gamma^{(2)} & -\mu^{(2)}-\gamma^{(2)} \end{bmatrix}$$

has state space $\{\{1,1\},\{1,2\},\{2,1\},\{2,2\}\}$.

We write $A \sim B$ and say that the two matrices are *stochastically equivalent* if there exists a permutation matrix $R$ such that $RAR = B$. The relation $\sim$ is an equivalence relation. Let

$$R = \begin{bmatrix} 1 & 0 & 0 & 0 \\ 0 & 0 & 1 & 0 \\ 0 & 1 & 0 & 0 \\ 0 & 0 & 0 & 1 \end{bmatrix},$$

then we see that $(\bigoplus_{k=1}^{2} \mathbb{Q}^{X|y_k} + \bigotimes_{k=1}^{2} \mathbb{Q}^{Y|x_k}) \sim (\bigoplus_{k=1}^{2} \mathbb{Q}^{Y|x_k} + \bigotimes_{k=1}^{2} \mathbb{Q}^{X|y_k})$. Checking the formulas describing the ordering from Proposition 2 for the $X$ process in the example above gives the states $\{1,2,1,2\}$ for $X$ in the first example and $\{1,1,2,2\}$ in the second one.

Note that this is only two of the total $(n^X n^Y)!$ possible configurations. One of the 22 other possible state spaces is

$$\{\{2,2\},\{1,1\},\{1,2\},\{2,1\}\}.$$

This ordering induce another permutation of the intensity matrix.

From now on, us assume that the process $Y$ is controlling the intensities of $X$ but $Y$ is locally independent of $X$. In other words $\mathbb{Q}^{Y|x_1} = \mathbb{Q}^{Y|x_2}$, with $\beta^{(1)} = \beta^{(2)}$ and $\gamma^{(1)} = \gamma^{(2)}$. Note that the processes in this example are for illustrative purposes only. In a more realistic setting the process $X$ would be observed and affected by the hidden process $Y$. Rydén (1996) does inference in the case when $X$ is a Poisson process and $Y$ is governing its intensity.

### 4. Measuring causality

In previous sections we introduced causality and local independence. This section develops a methodology testing for presence of causality. It is done by measuring a distance between the processes $X$ and $Y$. For this measure, we consider the Kullback–Leibler divergence.

#### 4.1. Kullback–Leibler Causality

Let $\mathbb{P}_0$ be a probability measure that is absolutely continuous with respect to $\mathbb{P}$. The Kullback–Leibler (KL) divergence is defined as $D_{\text{KL}}(\mathbb{P}_0 \| \mathbb{P}) = \mathbb{E}_{\mathbb{P}_0}\left[\log \frac{d\mathbb{P}_0}{d\mathbb{P}}\right]$. It can be viewed as the distance between two probability measures. Let $\mathbb{Q}_0$ and $\mathbb{Q}$ be intensity matrices coupled with Markov processes, and let $\mathbb{P}_{\mathbb{Q}_0}, \mathbb{P}_{\mathbb{Q}}$ be the probability measures parametrized by them. Let $A_T(i)$ denote the time spent in state $i$ during $[0,T]$ and denote $\mathbb{E}_{\mathbb{P}_{\mathbb{Q}_0}}^T [A_T(i)] = \mathfrak{A}_T$.

**Proposition 3.** *The Kullback–Leibler divergence for two continuous time Markov pro-*

cesses over $[0,T]$ parametrized by the intensity matrices $\mathbb{Q}_0$ and $\mathbb{Q}$ is

$$\mathrm{D}_{\mathrm{KL}}(\mathbb{P}_{\mathbb{Q}_0}\|\mathbb{P}_{\mathbb{Q}}) = \sum_i \mathfrak{A}_T(i)\left\{q_0(i\mid i) - q(i\mid i) + \sum_{j\neq i} q_0(j\mid i)\log\frac{q_0(j\mid i)}{q(j\mid i)}\right\},$$

with the convention $0\log\frac{0}{0} = 0$.

A corollary follows since $\mathfrak{A}_T$ is increasing in $T$: the KL divergence will either tend to infinity or to a constant as $T$ tends to infinity.

If the chain is not irreducible it is possible that the KL divergence converges to a constant. In that case there exists a closed communicating class such that that $q_0(i\mid i) - q(i\mid i) + \sum_{j\neq i} q_0(j\mid i)\log\frac{q_0(j\mid i)}{q(j\mid i)}$ is zero for all $i$ in that class.

Let $W = (X, Y)$ form a CTBN. Denote the probability which governs the network by $\mathbb{P}_{\mathbb{Q}^W}$. For a fixed value, $y_k$, the parameter $\mathbb{Q}^{X\mid Y=y_k}$ is not stochastic. Let $u$ be a uniform random variable on the interval $[0,T]$ and define $Y^{(T)} \triangleq Y_u$. Then the parameter $\mathbb{Q}^{X\mid Y^{(T)}}$ is a random variable in $Y^{(T)}$. It is natural to define $\mathbb{Q}^{X\mid\varnothing} \triangleq \mathbb{E}^T_{\mathbb{P}_{\mathbb{Q}^W}}[\mathbb{Q}^{X\mid Y^{(T)}}]$ and denote its elements by $q(x_j\mid x_i)$. Let $\mathbb{P}_{\mathbb{Q}^{X\mid\varnothing}}$ be a probability measure with parameter $\mathbb{Q}^{X\mid\varnothing}$ and let $\mathbb{P}_{\mathbb{Q}^{X\mid y_k}}$ be one which takes the component $Y$ into account. That is,

$$\mathbb{P}^T_{\mathbb{Q}^{X\mid\varnothing}}(X_{t+h} = x_j \mid X_t = x_i) \triangleq \sum_k \mathbb{P}_{\mathbb{Q}^{X\mid y_k}}(X_{t+h} = j \mid X_t = x_i)\frac{\mathfrak{A}^Y_T(y_k)}{T}, \text{ and}$$

$$\mathbb{P}_{\mathbb{Q}^{X\mid y_k}}(X_{t+h} = x_j \mid X_t = x_i) = \mathbb{P}_{\mathbb{Q}^W}(X_{t+h} = x_j, Y_{t+h} = y_k \mid X_t = x_i, Y_t = y_k).$$

Thus, by Doob (1953) and Lemma 1,

$$q(x_j\mid x_i) = \sum_k q(x_j\mid x_i, y_k)\frac{\mathfrak{A}^Y_T(y_k)}{T} = \begin{cases} \sum_k q(w^k_j\mid w^k_i), & i\neq j, \\ -\sum_{j\neq i} q(x_j\mid x_i), & i = j, \end{cases} \quad (9)$$

where $w^k_i = w_{i+n^X(k-1)}$ according to Proposition 2. The KL divergence requires absolute continuity between the two measures. We will define a causality measure based on this divergence so the following proposition is necessary.

**Proposition 4.** $\mathbb{P}_{\mathbb{Q}^{X\mid y_k}}$ is absolutely continuous with respect to $\mathbb{P}^T_{\mathbb{Q}^{X\mid\varnothing}}$. In general $\mathbb{P}^T_{\mathbb{Q}^{X\mid\varnothing}}$ is not absolutely continuous with respect to $\mathbb{P}_{\mathbb{Q}^{X\mid y_k}}$.

---

If the chain has no closed communicating class (or specifically: absorbing state), see Norris (1998), such that $q_0(i\mid i) - q(i\mid i) + \sum_{j\neq i} q_0(j\mid i)\log\frac{q_0(j\mid i)}{q(j\mid i)}$ is zero for all $i$ in that class (state) then the constant must be zero.

Define the amount of causality as

$$C_{\text{KL}}^T(X \leftarrow Y) \triangleq \mathbb{E}_{\mathbb{P}_{\mathbb{Q}^W}}^T [D_{\text{KL}}(\mathbb{P}_{\mathbb{Q}^{X|Y(T)}} \| \mathbb{P}_{\mathbb{Q}^{X|\varnothing}})].$$

This definition can be viewed as the expected distance over a interval $[0, T]$. Denote $\mathbb{E}_{\mathbb{P}_{\mathbb{Q}^W}}^T [A_T^Y(k)] = \mathfrak{A}_T^Y(k)$. Using this with Proposition 3 and Lemma 1 produces the following identity.

**Proposition 5.** *The causality for a CTBN is finite and may be written as*

$$C_{\text{KL}}^T(X \leftarrow Y) = \sum_k \frac{\mathfrak{A}_T^Y(y_k)}{T} D_{\text{KL}}(\mathbb{P}_{\mathbb{Q}^{X|y_k}} \| \mathbb{P}_{\mathbb{Q}^{X|\varnothing}}),$$

*where*

$$\begin{aligned}
D_{\text{KL}}(\mathbb{P}_{\mathbb{Q}^{X|y_k}} \| \mathbb{P}_{\mathbb{Q}^{X|\varnothing}}) &= \sum_i \mathfrak{A}_T^{X|y_k}(x_i) \bigg\{ q(x_i \mid x_i, y_k) \\
&\quad - q(x_i \mid x_i) + \sum_{j \neq i} q(x_j \mid x_i, y_k) \log \frac{q(x_j \mid x_i, y_k)}{q(x_j \mid x_i)} \bigg\} \\
&= \sum_i \mathfrak{A}_T^W(w_i^k) \bigg\{ q(x_i \mid x_i, y_k) \\
&\quad - q(x_i \mid x_i) + \sum_{j \neq i} q(w_j^k \mid w_i^k) \log \frac{q(w_j^k \mid w_i^k)}{q(x_j \mid x_i)} \bigg\}.
\end{aligned} \quad (10)$$

Here $\mathfrak{A}_T^{X|y_k}(x_i) = \int_0^T \mathbb{P}_{\mathbb{Q}^{X|y_k}}(X_t = x_i) \, dt$. It is more convenient to work with a "pure" process $W$. For instance, once the $W$ process is formed from observations $X$ and $Y$ it is easy to estimate $\mathfrak{A}_T^W(w_i^k) = \int_0^T \mathbb{P}_{\mathbb{Q}^{X|y_k}}(W_t = w_i^k)$, in contrast to $\mathfrak{A}_T^{X|y_k}(x_i)$ where the state of $Y$ needs to be taken into consideration.

4.1.1. *Average Causality and Calibration* The *average causality*

$$\frac{1}{T} C_{\text{KL}}^T(X \leftarrow Y)$$

is more convenient to work with. Since $A_T(i)$ is bounded above by $T$ for all $i$, so is the increase of the KL divergence. Therefore, we expect that for average causality

$$C_{\text{KL}}^1(X \leftarrow Y) \approx \frac{1}{T} C_{\text{KL}}^T(X \leftarrow Y) \approx \frac{1}{T+S} C_{\text{KL}}^{T+S}(X \leftarrow Y)$$

for any positive $T, S$ with equality in the last relation as $T$ tends to infinity.

**Proposition 6.** *The average causality $\frac{1}{T}\mathrm{C}_{\mathrm{KL}}^T(X \leftarrow Y)$ is bounded from above by the constant*

$$\max_{k,i} q(x_i \mid x_i, y_k) - q(x_i \mid x_i) + \sum_{j \neq i} q(x_j \mid x_i, y_k) \log \frac{q(x_j \mid x_i, y_k)}{q(x_j \mid x_i)}$$

*and the causality is bounded by $T$ scaled with the same constant.*

The KL calibration, $\kappa(x)$, is defined as the solution to the equation

$$x = \mathrm{D}_{\mathrm{KL}}(\mathrm{Be}(\tfrac{1}{2}) \| \mathrm{Be}(\kappa(x)))$$

and is given by $\kappa(x) = \frac{1}{2}(1 + \sqrt{1 + e^{-2x}})$, see McCulloch (1989). It is a standardized measure taking values in $[\frac{1}{2}, 1]$ and will be used to evaluate our results.

## 5. Tick data application

In this section an application in finance is studied. High resolution data with one data point for each time the price changes is under study. This data is called *tick-by-tick data*, or simply tick data. The observations are typically not equidistant in time. The second application is a simulation study on the example from Section 3.3 where causality is investigated.

An initial approach to modeling the tick-by-tick data is to use the *Skellam* process by Barndorff-Nielsen et al. (2012).

**Definition 1.** Let $N_t^+$ and $N_t^-$ be Poisson processes with intensities $\lambda^+$ and $\lambda^-$. Then the difference of these two processes is denoted

$$L_t = N_t^+ - N_t^-,$$

and is called a Skellam process.

The object of the section is to extend the Skellam model. Let $X$ and $Y$ be two independent processes taking values in the natural numbers. Define their difference process $Z_t = X_t - Y_t$. Since the CTBNs have finite state spaces an unbecoming technical condition, of no practical importance, is required. The processes must be bounded; therefore, let $\tau_M = \inf_t\{|X_t - Y_t| = M + 1\}$, and force $t$ to be in $[0, \tau_M]$.

**Proposition 7.** *Let the process $Z_t$ be the difference of two independent processes with independent increments. The process $Z$ is a composable process with components $(X,Y)$ if and only if the product of $\mathbb{P}(|X_{t+h} - X_t| > 0)$ and $\mathbb{P}(|Y_{t+h} - Y_t| > 0)$ is of order $o(h)$ for every $t < \tau_M$.*

*Proof.* A process with independent increments is a Markov process, see e.g. Doob (1953), thus

$$\mathbb{P}(X_{t+h} = k, Y_{t+h} = j \mid X_t \neq k, Y_t \neq j) = \mathbb{P}(|X_{t+h} - X_t| > 0)\mathbb{P}(|Y_{t+h} - Y_t| > 0).$$

In other words, $h^{-1}\mathbb{P}(X_{t+h} = k, Y_{t+h} = j \mid X_t \neq k, Y_t \neq j)$ tends to zero as $h$ tends to zero if and only if

$$\mathbb{P}(|X_{t+h} - X_t| > 0)\mathbb{P}(|Y_{t+h} - Y_t| > 0) \tag{11}$$

is of order $o(h)$.

Typically the processes $X$ and $Y$ will be *Lévy processes*. A stochastic process is said to be Lévy if it has independent and stationary increments and it is stochastically continuous. That is, if $\xi_t - \xi_s$ is independent of $\xi_t - \xi_r$ for any $t > s > r$ and the distribution of $\xi_{t+h} - \xi_t$ does not depend on $t$, and for every $\varepsilon > 0$ and for every $t$ the probability $\mathbb{P}(|\xi_{t+h} - \xi_t| > \varepsilon) = 0$ tends to zero as $h$ tends to zero.

For example, if $\xi$ is a Poisson process with intensity $\lambda$ then it satisfies

$$\mathbb{P}(|\xi_{t+h} - \xi_t| > \varepsilon) = \mathbb{P}(|\xi_{t+h} - \xi_t| > 0) = 1 - e^{-\lambda h} = \lambda h - o(h).$$

Evidently, it is stochastically continuous, and since it has independent and stationary increments it is Lévy. Its difference process, the Skellam process, is composable if the product in (11) is of sufficiently low order. Consider two Poisson processes with intensities $\lambda^+$ and $\lambda^-$ respectively. The product from (11) will be of order $(\lambda^+ h - o(h))(\lambda^- h - o(h)) = o(h)$. We deduce that the Skellam process is a composable process for $t < \tau_M$.

Summarizing, to verify that a difference process of two processes with independent increments is composable it suffices to check that the product in (11) is of $o(h)$ for every $t$.

**5.1. Model Proposal**

The price innovations are divided into upticks and downticks, i.e. positive and negative innovations respectively. Denote the price at time $t$ by $S_t$. Let $X_t$ be the increment of the price change and $Y_t$ the size of the decrease. Let $\tau_t$ denote the time for the most recent price change at time $t$. The difference process $Z_t$ is defined as

$$Z_t = X_t - Y_t = S_t - S_{\tau_t}$$

where $\tau_0 = \inf_s\{s : S_s \neq S_0\}$, $\tau_k = \inf_s\{s > \tau_{k-1} : S_s \neq S_{\tau_{k-1}}\}$, and $\tau_t = \max_k\{\tau_{k-1} < t\}$.

Thus, $Z_t$ is the difference of the price at time $t$ and of the previous price. The process moves in a finite space, thus $Z_t$ is a process taking only integer values. Since only price changes are considered, $Z_t$ can not be zero-valued. The process $X$ is constructed using the positive innovations of $Z$ while $Y$ is the value of the negative innovations in modulus. By definition, $X$ and $Y$ can not change value at the same time. Thus the process $W = (X, Y)$ forms a CTBN.

# 6. Results

We present inference done on the financial model in Section 5.1 and on the Markov modulated process from Section 3.3.

**6.1. Tick-by-tick EUR/USD**

The tick-by-tick data is modeled as the difference between two time homogeneous Markov processes following the model from Section 5.1. The data, price quotes of EUR/USD, is in millisecond resolution and contains roughly three hundred thousand data points spread out over three weeks. With millisecond resolution the process would have to be constant during long intervals in a standard time series framework. What is worse, there would be well over a billion data points for three weeks of data. This is gracefully evaded in the continuous time framework. The price data were originally retrieved from Hotspot FX, an electronic communication network. It provides a significant but not dominant part of the foreign exchange market liquidity, see Bech (2012), and King and Rime (2010) for details.

The task at hand is to estimate the intensity matrix for the process $W = (X, Y)$

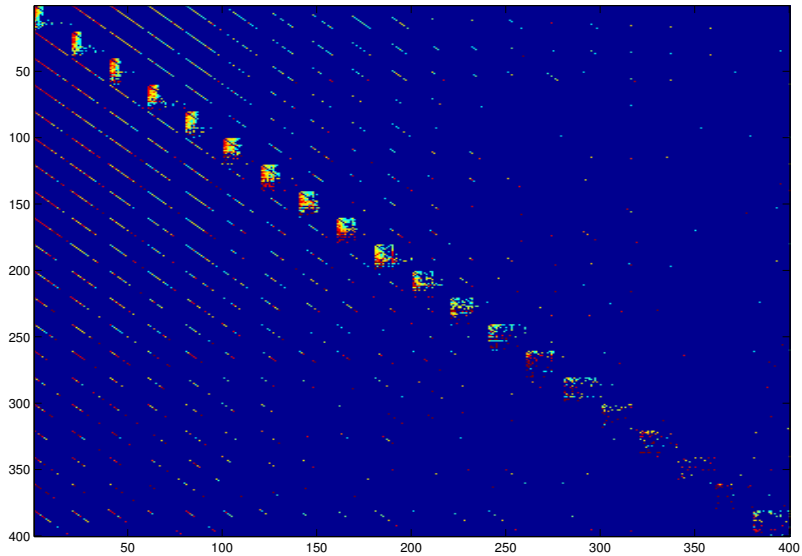

Figure 3: Intensity Matrix for EUR/USD data

using the price data. It is done by composing the processes and estimating the intensity matrix using formulas from Albert (1962), see Appendix B.

The result is presented as a heat map in Figure 3. In Figure 4 the intensities for $X$ when $Y$ is in state 1 and 10 respectively is seen. In the Skellam model the two matrices should be equal, this however does not seem to be the case. In the heatmap we allow the prices to jump 20 units. All larger movements will be capped at 20. The estimations indicates that the causality increases as the cap is lowered. Under the Skellam model only up or down movements are allowed, i.e. the cap is set at one. Under that assumption, the average causal KL calibration for the effect of $X$ on $Y$ is estimated to be 0.89. When a jump size of 5 is allowed it decreases to 0.78 and for 10 it is 0.67. The quantities are consistently smaller for the effect of $Y$ on $X$ but they are differing by less than one percent.

**6.2. Causality**

Recall the example with the processes $X$ and $Y$ from Section 3.3. In this section a simulation study of $Y$'s effect on $X$ is done. The empirical estimate of all parameters are denoted with a hat, for instance, the estimate of $C_{\mathrm{KL}}^T(X \leftarrow Y)$ is denoted as $\hat{C}_{\mathrm{KL}}^T(X \leftarrow Y)$. The estimators and their properties are given in Section B. The empirical estimate of causality is

$$\hat{C}_{\mathrm{KL}}^T(X \leftarrow Y) = \sum_{k=1}^{2} \hat{\mathfrak{A}}_T^Y k \sum_{i=1}^{2} \hat{\mathfrak{A}}_T^W(w_i^k) \Bigg\{ \hat{q}(x_i \mid x_i, y_k) \\ - \hat{q}(x_i \mid x_i) + \mathbb{I}(j \neq i) \hat{q}(w_j^k \mid w_i^k) \log \frac{\hat{q}(w_j^k \mid w_i^k)}{\hat{q}(x_j \mid x_i)} \Bigg\}.$$

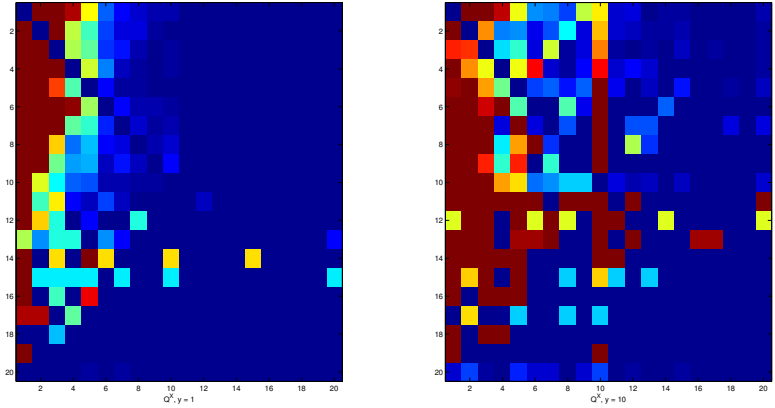

Figure 4: Comparison of submatrices

We set $T = 10^5$ and repeat the simulation a hundred thousand times producing estimates of the causality in Figure 5. The estimated causal effect of $X$ on $Y$ is close to zero with a very small variance. This means that our inference suggests that the process $X$ has no impact on $Y$. That is, there is evidence that $Y$ is locally independent of $X$. On the other hand, the same is not true for $X$. Instead the evidence support the belief $X \leftarrow Y$. Since $X$ is simulated with $Y$ as input, this is what we expected.

## 7. Discussion and future work

A natural extension of the applications is to look at processes with more than two components. For instance, investigating the causality between currency pairs. Another interesting aspect is to study what happens when a lagged version of the process is introduced. Will the process at time $t-1$ have a causal connection to that at time $t$? Medical data can be of high resolution and is an appealing alternative application.

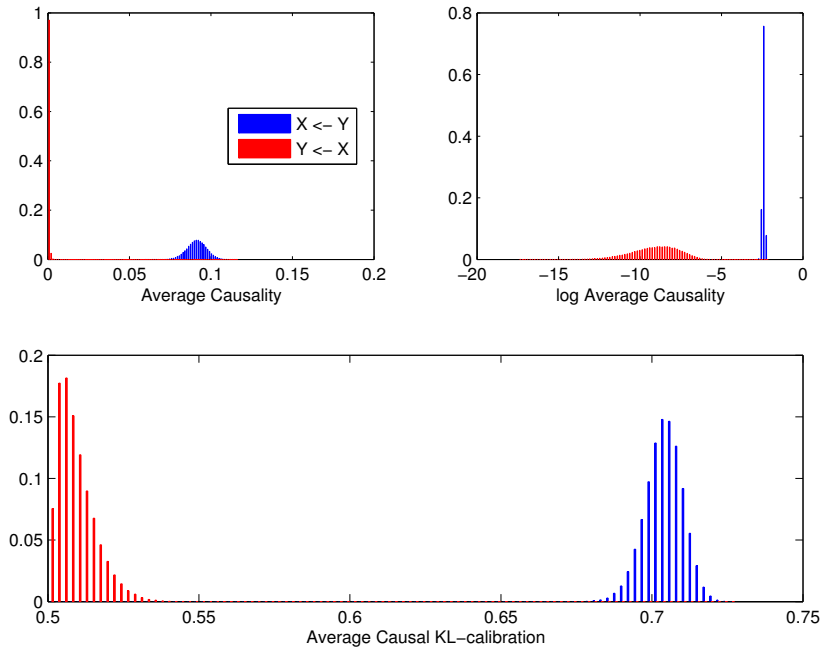

FIGURE 5: Causality estimates from the Markov modulated process

## Acknowledgements

The authors are grateful for the donated dataset with EUR/USD quotes. Funding for this research was provided by the Swedish Research Council (Grant Number 2009-5834).

## Appendix A. Proofs

**A.1. Proof of Lemma 1**

The composable property in (1), the probability that *both* elements change tends to zero as $h$ does the same, can be written as

$$\mathbb{P}(W_{t+h} = \{x_j, y_\ell\} \mid W_t = \{x_i, y_k\}) = o(h),$$

where $\lim_{h \to 0} o(h)/h = 0$. We will consider a chain in state $\{x_i, y_k\}$ moving to another state $\{x_j, y_\ell\}$ and three different cases: in the first case $x_j \neq x_i$, in the second $x_j = x_i$ and $y_\ell = y_k$; and in the third $x_j \neq x_i$, and $y_\ell \neq y_k$. Note that first case also covers $y_\ell \neq y_k$. Start with the case $x_j \neq x_i$ where

$$\mathbb{P}(X_{t+h} = x_j \mid X_t = x_i, Y_t = y_k) = \sum_{s=1}^{n^Y} \mathbb{P}(X_{t+h} = x_j, Y_{t+h} = y_s \mid X_t = x_i, Y_t = y_k)$$
$$= \mathbb{P}(X_{t+h} = x_j, Y_{t+h} = y_k \mid X_t = x_i, Y_t = y_k)$$
$$+ o(h),$$

holds by the composable property. Using this relation gives

$$\mathbb{P}(W_{t+h} = \{x_j, y_k\} \mid W_t = \{x_i, y_k\}) = \mathbb{P}(X_{t+h} = x_j, Y_{t+h} = y_k \mid X_t = x_i, Y_t = y_k)$$
$$= \mathbb{P}(X_{t+h} = x_j \mid X_t = x_i, Y_t = y_k) + o(h).$$

Dividing by $h$ and taking the limit yields

$$q(x_j, y_k \mid x_i, y_k) = q(x_j \mid x_i, y_k), \tag{12}$$

i.e. the element of the intensity matrix $\mathbb{Q}^W$ is given by one in $\mathbb{Q}^{X \mid y_k}$. In the same way, when $y_\ell \neq y_k$ we get $q(x_j, y_\ell \mid x_i, y_k) = q(y_\ell \mid x_i, y_k)$. For the second case, when the chain does not change, i.e. $x_j = x_i$ and $y_\ell = y_k$ we get

$$\mathbb{P}(X_{t+h} = x_i, Y_{t+h} = y_k \mid X_t = x_i, Y_t = y_k) = \mathbb{P}(X_{t+h} = x_i, \mid X_t = x_i, Y_t = y_k)$$
$$- \sum_{s \neq k} \mathbb{P}(X_{t+h} = x_i, Y_{t+h} = y_s \mid X_t = x_i, Y_t = y_k).$$

Dividing the above expression by $h$, taking the limit and using the relation in (12) produces the $q$-elements

$$q(x_i, y_k \mid x_i, y_k) = q(x_i \mid x_i, y_k) - \sum_{s \neq k} q(y_s \mid x_i, y_k) = q(x_i \mid x_i, y_k) + q(y_k \mid x_i, y_k).$$

Finally for the third case when $x_j \neq x_i$ and $y_\ell \neq y_k$ the composable property gives

$$\mathbb{P}(W_{t+h} = \{x_j, y_\ell\} \mid W_t = \{x_i, y_k\}) = o(h).$$

Thus $q(x_i, y_\ell \mid x_i, y_k) = 0$, and to summarize

$$q(x_j, y_\ell \mid x_i, y_k) = \begin{cases} q(x_j \mid x_i, y_k), & x_j \neq x_i, \\ q(y_\ell \mid x_i, y_k), & y_\ell \neq y_k, \\ q(x_i \mid x_i, y_k) + q(y_k \mid x_i, y_k), & x_j = x_i, y_\ell = y_k, \\ 0, & x_j \neq x_i, y_\ell \neq y_k. \end{cases}$$

**A.2. Proof of Proposition 1**

It is known by Lemma 1 that, each element in the $\mathbb{Q}^W$ matrix is a sum of elements from the $\mathbb{Q}^{X|y_k}$ and the $\mathbb{Q}^{Y|x_k}$ matrices. Thus we see that the contribution from $\mathbb{Q}^{X|y_k}$, corresponds to the block operator $\oplus$ from (4) while for $\mathbb{Q}^{Y|x_k}$ the contribution coincides with the expansion operator $\oslash$ defined in (5). In other words, the two operations expand the conditional matrices to the full space while keeping them consistent with the additive property from (3). This implies that the equality in Equation (6) holds for a composable process.

For the converse, that the operation produces a composable process, consider the operations $\oplus$ and $\oslash$. These operations produce matrices which when added together always satisfy the condition that only elements on the diagonal are changed. Thus

$$\mathbb{P}(\text{More than one process changes state in } [t, t+h])$$

is of order $o(h)$, which implies that the process is composable.

**A.3. Proof of Proposition 2**

The relation for $w$ follows from the construction of the operators. The established $w$ relation implies that

$$y = \frac{w-x}{n^X} + 1 = \frac{w}{n^X} + 1 - \frac{x}{n^X} = \frac{w}{n^X} + c,$$

where $c$ is a non-negative number not larger than one. Since $y$ only can take integer values it must be the smallest integer greater than or equal to $\frac{w}{n^X}$. This means that the constant $c$ can be ignored. For the final equality, observe that

$$x = w - n^X(y-1) = w - n^X\left(\left\lceil\frac{w}{n^X}\right\rceil - 1\right)$$
$$= 1 + (w-1) - n^X\left\lfloor\frac{w-1}{n^X}\right\rfloor = 1 + (w-1 \mod n^X)$$

where we use that $\lceil\frac{n}{m}\rceil - 1 = \lfloor\frac{n-1}{m}\rfloor$ when $n, m$ are positive integers.

## A.4. Proof of Proposition 3

Let $N_T(j \mid i)$ denote the number of times that the process transits from state $i$ to state $j$ during the time $[0, T]$. From Albert (1962) the log-likelihood for $\mathbb{Q}$ is given as

$$\ell_\mathbb{Q} = \log d\mathbb{P}_\mathbb{Q} = C + \sum_i \sum_{j \neq i} N_T(j \mid i) \log q(j \mid i) + \sum_i A_T(i) q(i \mid i).$$

Thus

$$\begin{aligned}\ell_{\mathbb{Q}_0} - \ell_\mathbb{Q} &= C - C + \sum_i \sum_{j \neq i} N_T(j \mid i) \log \frac{q_0(j \mid i)}{q(j \mid i)} \\ &\quad + \sum_i A_T(i)\bigl(q_0(i \mid i) - q(i \mid i)\bigr).\end{aligned} \quad (13)$$

Theorem 5.1 by Albert (1962) states that $\mathbb{E}^T_{\mathbb{P}_{\mathbb{Q}_0}}[N_T(j \mid i)] = q_0(j \mid i)\mathfrak{A}_T(i)$. Noting that $\mathrm{D}_{\mathrm{KL}}(\mathbb{P}_{\mathbb{Q}_0} \| \mathbb{P}_\mathbb{Q}) = \mathbb{E}_{\mathbb{P}_{\mathbb{Q}_0}}[\ell_{\mathbb{Q}_0} - \ell_\mathbb{Q}]$ and taking the expectation of (13) gives

$$\mathrm{D}_{\mathrm{KL}}(\mathbb{P}_{\mathbb{Q}_0} \| \mathbb{P}_\mathbb{Q}) = \sum_i \sum_{j \neq i} \mathfrak{A}_T(i) q_0(j \mid i) \log \frac{q_0(j \mid i)}{q(j \mid i)} + \sum_i \mathfrak{A}_T(i)\bigl(q_0(i \mid i) - q(i \mid i)\bigr).$$

Finally, collecting the factors gives the result.

## A.5. Proof of Proposition 4

Theorem 3.3 by Albert (1962) states: if $q(x_j \mid x_i, y_k)$ vanishes whenever $q(x_j \mid x_i)$ does, then the probability measure on $(\mathsf{W}, \mathbb{P})$ under $\mathbb{Q}^{X \mid y_k}$ is absolutely continuous with respect to the one under $\mathbb{Q}^{X \mid \varnothing}$.

If $q(x_j \mid x_i, y_k)$ is positive then $X$ can transit from $i$ to $j$, thus the probability of such a move must be positive. This occurs only if $q(x_j \mid x_i)$ is positive since there is a direct relation between that probability and $q(x_j \mid x_i)$, see e.g. Doob (1953).

By the same theorem, if any $q(x_j \mid x_i, y_k)$ is equal to zero while $q(x_j \mid x_i)$ is strictly positive, then $\mathbb{P}^T_{\mathbb{Q}^{X \mid \varnothing}}$ is not absolutely continuous with respect to $\mathbb{P}_{\mathbb{Q}^{X \mid y_k}}$.

It is possible that the process $X$ can move from $i$ to $j$ under only some states of $Y$. That is, $q(x_j \mid x_i, y_k)$ can be zero-valued for some $k$ while $q(x_j \mid x_i) > 0$ implying that $\mathbb{P}^T_{\mathbb{Q}^{X\mid\varnothing}}$ is not absolutely continuous with respect to $\mathbb{P}_{\mathbb{Q}^{X\mid y_k}}$.

### A.6. Proof Proposition 5

Firstly, let $\phi$ be a function such that $\phi(Y)$ has a finite expectation under $\mathbb{P}_{\mathbb{Q}^W}$, and let $u$ be a uniform random variable in $[0, T]$ then

$$\mathbb{E}^T_{\mathbb{P}_{\mathbb{Q}^W}}[\phi(Y)] = \sum_k \phi(y_k)\mathbb{E}[\mathbb{P}_{\mathbb{Q}^W}(Y_u = y_k)] = \sum_k \phi(y_k)\frac{1}{T}\int_0^T \mathbb{P}_{\mathbb{Q}^W}(Y_t = y_k)]\,dt$$
$$= \sum_k \phi(y_k)\frac{\mathfrak{A}^Y_T(y_k)}{T}.$$

Instead of $\phi$, use the Kullback–Leibler divergence to obtain

$$\sum_k D_{\mathrm{KL}}(\mathbb{P}_{\mathbb{Q}^{X\mid y_k}} \| \mathbb{P}_{\mathbb{Q}^{X\mid\varnothing}})\frac{\mathfrak{A}^Y_T(y_k)}{T}.$$

Proposition 4 states that the measures are absolutely continuous as required for the KL divergence to be finite. The KL divergence for two probability measures parametrized by intensity matrices are given by Proposition 3 and Lemma 1 delivers the elements of the intensity matrix. Thus

$$C^T_{\mathrm{KL}}(X \leftarrow Y) = \sum_k \frac{\mathfrak{A}^Y_T(y_k)}{T}\sum_i \mathfrak{A}^{X\mid y_k}_T(x_i)\bigg\{q(x_i \mid x_i, y_k)$$
$$- q(x_i \mid x_i) + \sum_{j \neq i} q(x_j \mid x_i, y_k)\log\frac{q(x_j \mid x_i, y_k)}{q(x_j \mid x_i)}\bigg\}.$$

Finally, by using Proposition 2 we find that $\mathfrak{A}^{X\mid y_k}_T(x_i) = \mathfrak{A}^W_T(w^k_i)$ and analogously

$$q(x_i \mid x_i, y_k) + \sum_{j \neq i} q(x_j \mid x_i, y_k)\log q(x_j \mid x_i, y_k)$$
$$= q(x_i \mid x_i, y_k) + \sum_{j \neq i} q(w^k_j \mid w^k_i)\log q(w^k_j \mid w^k_i)$$

which concludes the proof.

### A.7. Proof of Proposition 6

Let $c_{ki}$ denote

$$q(x_i \mid x_i, y_k) - q(x_i \mid x_i) + \sum_{j \neq i} q(x_j \mid x_i, y_k)\log\frac{q(x_j \mid x_i, y_k)}{q(x_j \mid x_i)},$$

and $c^*$ the maximum over $k, i$. Then

$$\mathrm{D}_{\mathrm{KL}}(\mathbb{P}_{\mathbb{Q}^{X|y_k}} \| \mathbb{P}_{\mathbb{Q}^{X|\varnothing}}) = \sum_i \mathfrak{A}_T^{X|y_k}(x_i) c_{ki} \le c^* \sum_i \mathfrak{A}_T^{X|y_k}(x_i) = c^* T.$$

Thus $\mathrm{C}_{\mathrm{KL}}^T(X \leftarrow Y) \le c^* T \sum_k \frac{\mathfrak{A}_T^Y(y_k)}{T} = c^* T$ and we deduce that the average causality is bounded from above by $c^*$.

## Appendix B. Estimators

This section closely follows Albert (1962) and Basawa and Rao (1980). It is included for completeness. Let $W$ be a time homogeneous continuous time finite Markov process, and let the elements of its intensity matrix be denoted by $q(j \mid i)$. Let

$$N_T^{(m)}(j \mid i) = \begin{array}{c} \text{the total number of transitions from} \\ \text{state } i \text{ to state } j \text{ observed during } m \text{ trials, and} \end{array}$$

$$A_T^{(m)}(i) = \begin{array}{c} \text{the total length of time that} \\ \text{state } i \text{ is occupied during } m \text{ trials.} \end{array}$$

The likelihood for $\mathbb{Q}$ is then

$$\log L_{\mathbb{Q}}^{(m)} = C_m + \sum_i \sum_j N_T^{(m)}(j \mid i) \log q(j \mid i) + \sum_{i=1}^M A_T^{(m)}(i) q(i \mid i),$$

where $C_m$ is finite almost surely and does not depend on $\mathbb{Q}$. The maximum-likelihood estimate is given by $\hat{q}_T^{(m)}(j \mid i) = \frac{N_T^{(m)}(j|i)}{A_T^{(m)}(i)}$, where we use the convention $\hat{q}_T^{(m)}(j \mid i) = 0$ if $i \ne j$ and $A_T^{(m)}(i) = 0$.

### B.1. Estimates for CTBNs

As before, let $A_T(i)$ be the time a process spends in state $i$ and $N_T(i \mid j)$ the number of time a process transits from $i$ to $j$ during the interval $[0, T]$.

The likelihood estimate of the elements in $\mathbb{Q}^{X|y_k}$ are given by

$$\hat{q}(x_j \mid x_i, y_k) = \frac{N_T^{X|y_k}(x_j \mid x_i)}{A_T^{X|y_k}(x_j)}$$

with the convention $\hat{q}(x_j \mid x_i, y_k) = 0$ if $A_T^{X|y_k}(x_j) = 0$, i.e. if the process never enters state $x_j$ during the time $[0, T]$. The estimate of

$$\frac{1}{T} \mathbb{E}_{\mathbb{P}_{\mathbb{Q}^W}}^T [A_T^Y(y_k)] = \frac{\mathfrak{A}_T^Y(y_k)}{T}$$

is given by

$$\frac{1}{T}A_T^Y(y_k) = \frac{1}{T}\hat{\mathfrak{A}}_T^Y(y_k) = \frac{1}{T}\int_0^T \mathbb{I}(Y_t = y_k)\,dt$$

and the expectation of the integral is $\frac{1}{T}\int_0^T \mathbb{P}(Y_t = k)\,dt$ so the estimator is unbiased. Further, for any indicator variable

$$\begin{aligned}
\mathbb{V}\Big[\int_0^T \mathbb{I}(\cdot_t)\,dt\Big] &= \mathbb{E}\Big[\Big(\int_0^T \mathbb{I}(\cdot_t)\,dt\Big)^2\Big] - \mathbb{E}\Big[\Big(\int_0^T \mathbb{I}(\cdot_t)\,dt\Big)\Big]^2 \\
&\leq \mathbb{E}\Big[\int_0^T \mathbb{I}(\cdot_t)^2\,dt\Big] - \mathbb{E}\Big[\Big(\int_0^T \mathbb{I}(\cdot_t)\,dt\Big)\Big]^2 \\
&= \mathbb{E}\Big[\int_0^T \mathbb{I}(\cdot_t)\,dt\Big]\Big(1 - \mathbb{E}\Big[\int_0^T \mathbb{I}(\cdot_t)\,dt\Big]\Big) \leq \frac{T}{2},
\end{aligned}$$

where we use Jensen's inequality for the first inequality. In the second we use that $f(T)(1-f(T)) = T^2\frac{f(T)}{T}(\frac{1}{T} - \frac{f(T)}{T})$ is maximized at $f(T) = \frac{T}{2}$ for functions $f$ taking values in $[0,T]$. Thus $\frac{1}{T^2}\mathbb{V}_{\mathbb{P}_{\mathbb{Q}^W}}^T[A_T^Y(y_k)] \leq \frac{1}{2T}$ resulting in the following proposition.

**Proposition 8.** $\mathbb{E}_{\mathbb{P}_{\mathbb{Q}^W}}^T[\frac{1}{T}A_T^Y(y_k)] = \frac{\mathfrak{A}_T^Y(y_k)}{T}$ and $\mathbb{V}_{\mathbb{P}_{\mathbb{Q}^W}}^T[\frac{1}{T}A_T^Y] \leq \frac{1}{2T}$.

The elements in the estimator of $\mathbb{Q}^{X|\varnothing}$ are denoted $\hat{q}(x_i \mid x_j)$ and given by

$$\hat{q}(x_j \mid x_i) = \frac{N_T^X(x_j \mid x_i)}{A_T^X(x_j)}.$$

The estimate does not take $Y$ into account at all which is what we want. However the quantity in $\mathbb{Q}^{X|\varnothing}$ must consider $Y$ in its definition since the transitions of $X$ are not available without $Y$. Plug the estimate of $q$ into the definition of $\mathbb{Q}^{X|\varnothing}$ from (9) to obtain

$$\begin{aligned}
&\mathbb{E}_{\mathbb{P}_{\mathbb{Q}^W}}^T\Big[\sum_k \hat{q}(x_j \mid x_i, y_k)\frac{\hat{\mathfrak{A}}_T^Y(y_k)}{T}\Big] \\
&= \sum_k \mathbb{E}_{\mathbb{P}_{\mathbb{Q}^W}}^T[\hat{q}(x_j \mid x_i, y_k)]\mathbb{E}_{\mathbb{P}_{\mathbb{Q}^W}}^T\Big[\frac{1}{T}\int_0^T \mathbb{I}(Y_t = y_k)\,dt\Big] \\
&= \sum_k \frac{1}{T}\int_0^T q(x_j \mid x_i, y_k)\mathbb{P}_{\mathbb{Q}^W}(Y_t = y_k)\,dt \\
&= \frac{1}{T}\int_0^T \lim_{h\to 0}\frac{1}{h}\sum_k \mathbb{P}_{\mathbb{Q}^W}(X_{t+h} = x_j, Y_t = y_k \mid X_t = x_i)\,dt \\
&= \frac{1}{T}\int_0^T \lim_{h\to 0}\frac{1}{h}\mathbb{P}_{\mathbb{Q}^W}(X_{t+h} = x_j \mid X_t = x_i)\,dt \\
&= q(x_j \mid x_i)\frac{1}{T}\int_0^T dt = q(x_j \mid x_i) = \mathbb{E}_{\mathbb{P}_{\mathbb{Q}^W}}^T\Big[\frac{N_T^X(x_j \mid x_i)}{A_T^X(x_j)}\Big],
\end{aligned}$$

where we use that the estimator $\hat{q}(x_j \mid x_i, y_k) = \frac{N_T^{X|y_k}(x_j|x_i)}{A_T^{X|y_k}(x_j)}$ is independent of $\frac{1}{T}\int_0^T \mathbb{I}(Y_t = k)\,dt$. This shows that the definition is consistent with the estimator we want to use.

Note that while $\mathbb{P}_{\mathbb{Q}^W}(X_{t+h} = x_j \mid X_t = x_i)$ coincides with

$$\mathbb{P}_{\mathbb{Q}^{X|\varnothing}}^T(X_{t+h} = x_j \mid X_t = x_i)$$

the two are not the same measure. Under $\mathbb{P}_{\mathbb{Q}^{X|\varnothing}}^T$ the process $X$ is locally independent of $Y$ and for instance

$$\mathbb{P}_{\mathbb{Q}^{X|\varnothing}}^T(X_{t+h} = x_j \mid X_t = x_i, Y_t = y_k) = \mathbb{P}_{\mathbb{Q}^{X|\varnothing}}^T(X_{t+h} = x_j \mid X_t = x_i),$$

which is not true under $\mathbb{P}_{\mathbb{Q}^W}$.

## Appendix C. Some notes for nonfinite sample spaces

Let $X$ and $Y$ have measurable spaces $(\mathsf{X}, \mathcal{X})$ and $(\mathsf{Y}, \mathcal{Y})$ respectively where $\mathsf{X}$ and $\mathsf{Y}$ are subsets of the real line and $\mathcal{X}, \mathcal{Y}$ their corresponding Borel sigma-algebras. Then define a process to be composable if

$$\mathbb{P}(\{|X_{t+h} - X_t| > \varepsilon \cap |Y_{t+h} - Y_t| > \varepsilon \mid X_t = x_i, Y_t = y_k) = o(h), \tag{14}$$

for every $\varepsilon > 0$ and $t \geq 0$. The calculations from Lemma 1 can be repeated to produce

$$\begin{aligned}&\mathbb{P}(X_{t+h} \in A \mid X_t = x_i, Y_t = y_k)\\ &= \int_{\mathsf{Y}} \mathbb{P}(X_{t+h} \in A, Y_{t+h} \in du \mid X_t = x_i, Y_t = y_k)\\ &= \mathbb{P}(X_{t+h} \in A, Y_{t+h} \in Y_t \pm \varepsilon \mid X_t = x_i, Y_t = y_k) + o(h),\end{aligned} \tag{15}$$

where $A = \mathsf{X} \setminus \{x_k \pm \varepsilon\}$. Define $B$ in the same way for $Y$: $B = \mathsf{Y} \setminus \{y_k \pm \varepsilon\}$. Dividing by $h$ and taking the limit yields

$$\begin{aligned}q(A, y_k \pm \varepsilon \mid x_i, y_k) &= q_X(A \mid x_i, y_k), A \in \mathcal{X}\\ q(x_k \pm \varepsilon, B \mid x_i, y_k) &= q_Y(B \mid x_i, y_k), B \in \mathcal{Y}.\end{aligned} \tag{16}$$

For the case when both processes change we have

$$\mathbb{P}(X_{t+h} \in A, Y_{t+h} \in B \mid X_t = x_i, Y_t = y_k) = o(h) \tag{17}$$

by definition since the processes are composable. For the final case, let $A^C$ denote the complement of the set $A$,

$$\mathbb{P}(X_{t+h} \in A^C, Y_{t+h} \in B^C \mid X_t = x_i, Y_t = y_k), \tag{18}$$

we use

$$\begin{aligned}
&\mathbb{P}(X_{t+h} \in A^C \mid X_t = x_i, Y_t = y_k) \\
&= \mathbb{P}(X_{t+h} \in A^C, Y_{t+h} \in \mathsf{Y} \mid X_t = x_i, Y_t = y_k) \\
&= \mathbb{P}(X_{t+h} \in A^C, Y_{t+h} \in B \mid X_t = x_i, Y_t = y_k) \\
&\quad + \mathbb{P}(X_{t+h} \in A^C, Y_{t+h} \in B^C \mid X_t = x_i, Y_t = y_k).
\end{aligned} \tag{19}$$

Thus

$$\begin{aligned}
&\mathbb{P}(X_{t+h} \in A^C, Y_{t+h} \in B^C \mid X_t = x_i, Y_t = y_k) \\
&= \mathbb{P}(X_{t+h} \in A^C \mid X_t = x_i, Y_t = y_k) \\
&\quad - \mathbb{P}(X_{t+h} \in A^C, Y_{t+h} \in B \mid X_t = x_i, Y_t = y_k) \\
&= \mathbb{P}(X_{t+h} \in A^C \mid X_t = x_i, Y_t = y_k) \\
&\quad - \mathbb{P}(Y_{t+h} \in B \mid X_t = x_i, Y_t = y_k) + o(h).
\end{aligned} \tag{20}$$

Dividing by $h$ and taking the limit gives us that

$$\begin{aligned}
q(A^C, B^C \mid x_i, y_k) &= q_X(A^c \mid x_i, y_k) - q_Y(B \mid x_i, y_k) \\
&= q_Y(B^c \mid x_i, y_k) - q_X(A \mid x_i, y_k) = q_Y(B^c \mid x_i, y_k) + q_X(A^C \mid x_i, y_k).
\end{aligned}$$

Finally we see that this definition is consistent with the one from Equation (1) in the introduction. Note that if the condition that the space should be finite is removed the relation still holds. It was not mentioned in Lemma 1 but the equations derived do not require the state space to be finite.

*References*

Odd O Aalen, Kjetil Røysland, Jon Michael Gran, and Bruno Ledergerber. Causality, mediation and time: a dynamic viewpoint. *Journal of the Royal Statistical Society: Series A (Statistics in Society)*, 175(4):831–861, 2012.

Arthur Albert. Estimating the infinitesimal generator of a continuous time, finite state Markov process. *The Annals of Mathematical Statistics*, 33(2):727–753, 1962.

Pierre-Olivier Amblard and Olivier JJ Michel. The relation between granger causality and directed information theory: a review. *Entropy*, 15(1):113–143, 2012.

Ole E Barndorff-Nielsen, David G Pollard, and Neil Shephard. Integer-valued Lévy processes and low latency financial econometrics. *Quantitative Finance*, 12(4):587–605, 2012.

Ishwar V Basawa and BLS Prakasa Rao. *Statistical inference for stochastic processes*. Academic press London, 1980.

Morten Bech. FX volume during the financial crisis and now. *BIS Quarterly Review*, 1:33–43, 2012.

Vanessa Didelez. Graphical models for composable finite Markov processes. *Scandinavian Journal of Statistics*, 34(1):169–185, 2007.

Joseph L Doob. *Stochastic processes*, volume 101. New York Wiley, 1953.

Michael Eichler. Graphical modelling of multivariate time series. *Probability Theory and Related Fields*, 153(1-2):233–268, 2012.

Jean-Pierre Florens and Denis Fougere. Noncausality in continuous time. *Econometrica: Journal of the Econometric Society*, pages 1195–1212, 1996.

Anne Gégout-Petit and Daniel Commenges. A general definition of influence between stochastic processes. *Lifetime data analysis*, 16(1):33–44, 2010.

Christian Gourieroux, Alain Monfort, and Eric Renault. Kullback causality measures. *Annales d'Economie et de Statistique*, pages 369–410, 1987.

Clive WJ Granger. Investigating causal relations by econometric models and cross-spectral methods. *Econometrica: Journal of the Econometric Society*, pages 424–438, 1969.

Katerina Hlaváčková-Schindler, Milan Paluš, Martin Vejmelka, and Joydeep Bhattacharya. Causality detection based on information-theoretic approaches in time series analysis. *Physics Reports*, 441(1):1–46, 2007.

Jan M Hoem. Purged and partial Markov chains. *Scandinavian Actuarial Journal*, 1969(3-4):147–155, 1969.

A Kaiser and T Schreiber. Information transfer in continuous processes. *Physica D: Nonlinear Phenomena*, 166(1):43–62, 2002.

Michael R King and Dagfinn Rime. The $4 trillion question: what explains FX growth since the 2007 survey? *BIS Quarterly Review*, 4:27–42, 2010.

Robert E McCulloch. Local model influence. *Journal of the American Statistical Association*, 84(406):473–478, 1989.

Uri Nodelman, Christian R Shelton, and Daphne Koller. Continuous time Bayesian networks. In *Proceedings of the Eighteenth conference on Uncertainty in artificial intelligence*, pages 378–387. Morgan Kaufmann Publishers Inc., 2002.

James R Norris. *Markov chains*. Number 2008. Cambridge university press, 1998.

Tobias Rydén. An EM algorithm for estimation in Markov-modulated Poisson processes. *Computational Statistics & Data Analysis*, 21(4):431–447, 1996.

Tore Schweder. Composable Markov processes. *Journal of applied probability*, 7(2): 400–410, 1970.

Abd-Krim Seghouane and Shun-ichi Amari. Identification of directed influence: Granger causality, kullback-leibler divergence, and complexity. *Neural computation*, 24(7):1722–1739, 2012.

George Sugihara, Robert May, Hao Ye, Chih-hao Hsieh, Ethan Deyle, Michael Fogarty, and Stephan Munch. Detecting causality in complex ecosystems. *science*, 338(6106): 496–500, 2012.

Tsachy Weissman, Young-Han Kim, and Haim H Permuter. Directed information, causal estimation, and communication in continuous time. *Information Theory, IEEE Transactions on*, 59(3):1271–1287, 2013.